\pdfoutput=1
\documentclass[letterpaper, 10 pt, conference]{ieeeconf}
\IEEEoverridecommandlockouts     

\usepackage{lipsum}

\usepackage{multibib}
\newcites{app}{Appendix References}

\usepackage{wrapfig}
\usepackage{xcolor} 

\usepackage{amsmath}
\usepackage{amsfonts}
\usepackage{amsthm}
\usepackage{amssymb}
\usepackage{mathrsfs}  
\usepackage{bm}
\usepackage[ruled,vlined]{algorithm2e}
\usepackage{mathdots} 
\usepackage{dsfont} 
\usepackage{accents} 
\usepackage{tikz-cd} 
\usepackage{adjustbox} 
\usepackage{mathtools} 

\usepackage{caption}
\usepackage{multirow}
\usepackage{multicol}
\usepackage{mdframed} 

\usepackage{listings}
\usepackage{graphicx}

\usepackage{balance}

\newcommand{\comment}[1]{} 
 
\newcommand{\norm}[1]{\left\lVert#1\right\rVert} 
\newcommand\mat[1]{\begin{bmatrix}#1\end{bmatrix}} 
\newcommand\eqs[1]{\begin{equation}\begin{split}#1\end{split}\end{equation}} 
\newcommand\eqsnn[1]{\begin{equation*}\begin{split}#1\end{split}\end{equation*}} 
\newcommand\inner[2]{\left\langle#1,#2\right\rangle} 

\DeclareMathOperator*{\diag}{diag} 

\DeclareMathOperator*{\conv}{conv} 
\DeclareMathOperator*{\maximize}{maximize}
\DeclareMathOperator*{\subto}{subject\;to}
\DeclareMathOperator*{\erf}{erf} 
\newcommand\braces[1]{\left\{#1\right\}} 
\newcommand\brackets[1]{\left[#1\right]} 
\newcommand\parens[1]{\left(#1\right)} 
\newcommand\abs[1]{\left|#1\right|} 
\newcommand\wedgeop[1]{\left[#1\right]_\times}
\newcommand\R{\mathbb{R}} 

\newcommand{\pdf}{p} 
\newcommand{\pos}{x} 
\newcommand{\posSet}{\mathcal{X}}

\newtheorem{theorem}{Theorem}
\newtheorem{lemma}{Lemma}

\newtheorem*{definition*}{Definition}

\newtheorem{proposition}{Proposition}

\usepackage[bookmarks=true, hidelinks]{hyperref}

\makeatletter
\let\save@mathaccent\mathaccent
\newcommand*\if@single[3]{%
  \setbox0\hbox{${\mathaccent"0362{#1}}^H$}%
  \setbox2\hbox{${\mathaccent"0362{\kern0pt#1}}^H$}%
  \ifdim\ht0=\ht2 #3\else #2\fi
  }
\newcommand*\rel@kern[1]{\kern#1\dimexpr\macc@kerna}
\newcommand*\widebar[1]{\@ifnextchar^{{\wide@bar{#1}{0}}}{\wide@bar{#1}{1}}}
\newcommand*\wide@bar[2]{\if@single{#1}{\wide@bar@{#1}{#2}{1}}{\wide@bar@{#1}{#2}{2}}}
\newcommand*\wide@bar@[3]{%
  \begingroup
  \def\mathaccent##1##2{%
    \let\mathaccent\save@mathaccent
    \if#32 \let\macc@nucleus\first@char \fi
    \setbox\z@\hbox{$\macc@style{\macc@nucleus}_{}$}%
    \setbox\tw@\hbox{$\macc@style{\macc@nucleus}{}_{}$}%
    \dimen@\wd\tw@
    \advance\dimen@-\wd\z@
    \divide\dimen@ 3
    \@tempdima\wd\tw@
    \advance\@tempdima-\scriptspace
    \divide\@tempdima 10
    \advance\dimen@-\@tempdima
    \ifdim\dimen@>\z@ \dimen@0pt\fi
    \rel@kern{0.6}\kern-\dimen@
    \if#31
      \overline{\rel@kern{-0.6}\kern\dimen@\macc@nucleus\rel@kern{0.4}\kern\dimen@}%
      \advance\dimen@0.4\dimexpr\macc@kerna
      \let\final@kern#2%
      \ifdim\dimen@<\z@ \let\final@kern1\fi
      \if\final@kern1 \kern-\dimen@\fi
    \else
      \overline{\rel@kern{-0.6}\kern\dimen@#1}%
    \fi
  }%
  \macc@depth\@ne
  \let\math@bgroup\@empty \let\math@egroup\macc@set@skewchar
  \mathsurround\z@ \frozen@everymath{\mathgroup\macc@group\relax}%
  \macc@set@skewchar\relax
  \let\mathaccentV\macc@nested@a
  \if#31
    \macc@nested@a\relax111{#1}%
  \else
    \def\gobble@till@marker##1\endmarker{}%
    \futurelet\first@char\gobble@till@marker#1\endmarker
    \ifcat\noexpand\first@char A\else
      \def\first@char{}%
    \fi
    \macc@nested@a\relax111{\first@char}%
  \fi
  \endgroup
}
\makeatother

\begin{document}

\title{
    PONG: \underline{P}robabilistic \underline{O}bject \underline{N}ormals for \underline{G}rasping \\ via Analytic Bounds on Force Closure Probability
}

\author{
   Albert H. Li$^\dagger$, Preston Culbertson$^\ddagger$, Aaron D. Ames$^{\dagger,\ddagger}$%
       \thanks{$\dagger$ A. H. Li and A. D. Ames are with the Department of Computing and Mathematical Sciences, California Institute of Technology, Pasadena, CA 91125, USA, \texttt{\{alberthli, ames\}@caltech.edu}.}%
        \thanks{$\ddagger$ P. Culbertson and A. D. Ames are with the Department of Civil and Mechanical Engineering, California Institute of Technology, Pasadena, CA 91125, USA, \texttt{\{pculbert, ames\}@caltech.edu}.}%
}


\maketitle

\begin{abstract}
Classical approaches to grasp planning are deterministic, requiring perfect knowledge of an object's pose and geometry. In response, data-driven approaches have emerged that plan grasps entirely from sensory data. While these data-driven methods have excelled in generating parallel-jaw and power grasps, their application to precision grasps (those using the fingertips of a dexterous hand, e.g, for tool use) remains limited. Precision grasping poses a unique challenge due to its sensitivity to object geometry, which allows small uncertainties in the object's shape and pose to cause an otherwise robust grasp to fail. In response to these challenges, we introduce Probabilistic Object Normals for Grasping (PONG), a novel, analytic approach for calculating a conservative estimate of force closure probability in the case when contact locations are known but surface normals are uncertain. We then present a practical application where we use PONG as a grasp metric for generating robust grasps both in simulation and real-world hardware experiments. Our results demonstrate that maximizing PONG efficiently produces robust grasps, even for challenging object geometries, and that it can serve as a well-calibrated, uncertainty-aware metric of grasp quality.
\end{abstract}

\maketitle

\section{Introduction}\label{sec:intro}
Grasp synthesis has been a canonical problem in robotic manipulation since the field's inception. Despite decades of work, dexterous grasps are still challenging to synthesize, since multifinger hands have complex kinematics and high-dimensional grasp parameterizations. Broadly speaking, two types of approaches toward dexterous grasp synthesis exist.
\textit{Analytic} approaches evaluate grasp quality using a \textit{metric} \cite{roa2014_graspmetricssurvey} and then maximize it using any optimization technique. Though analytic methods offer principled guarantees, they suffer from two problems: (i) they usually assume perfect knowledge of an object's geometry, and (ii) they are typically hard to optimize efficiently while enforcing kinematic and collision constraints \cite{li1988_taskorientedgrasping}.
In response, many \textit{learning-based} methods account for uncertainty with data but lack guarantees, often checking constraints are satisfied \textit{post hoc} via rejection sampling rather than enforcing them during synthesis \cite{kappler2015_bigdatagrasping, aktas2019_deepdexterousgrasping, shao2019_unigrasp}. To ensure sample efficiency, large amounts of high-quality data are required, which may be difficult to generate or collect.

Ideally, a grasp synthesis method should be uncertainty-aware, computationally-efficient, and generalize to a broad class of object geometries. To that end, our contributions are as follows. First, we develop a novel analytic theory of probabilistic force closure, PONG: \underline{\textbf{P}}robabilistic \underline{\textbf{O}}bject \underline{\textbf{N}}ormals for \underline{\textbf{G}}rasping. Given known contact locations and a model of surface normal uncertainty, PONG computes a lower bound on the probability that a grasp is force closure, which we maximize to synthesize uncertainty-aware precision grasps. We demonstrate PONG's effectiveness by using it as a curvature-aware grasp metric in both simulation\footnote{Open-source implementation available at \href{https://github.com/alberthli/pong}{github.com/alberthli/pong}.} and hardware experiments, where we treat the object's curvature as a proxy for uncertainty about its surface geometry. We show in simulation that, as PONG increases, the failure rate in grasps decreases dramatically. In hardware, even under challenging real-world conditions and using approximate, learning-based object models constructed only from visual data, we can achieve a nearly 70\% grasp success rate.

\begin{figure}
    \centering
    \includegraphics[width=\linewidth]{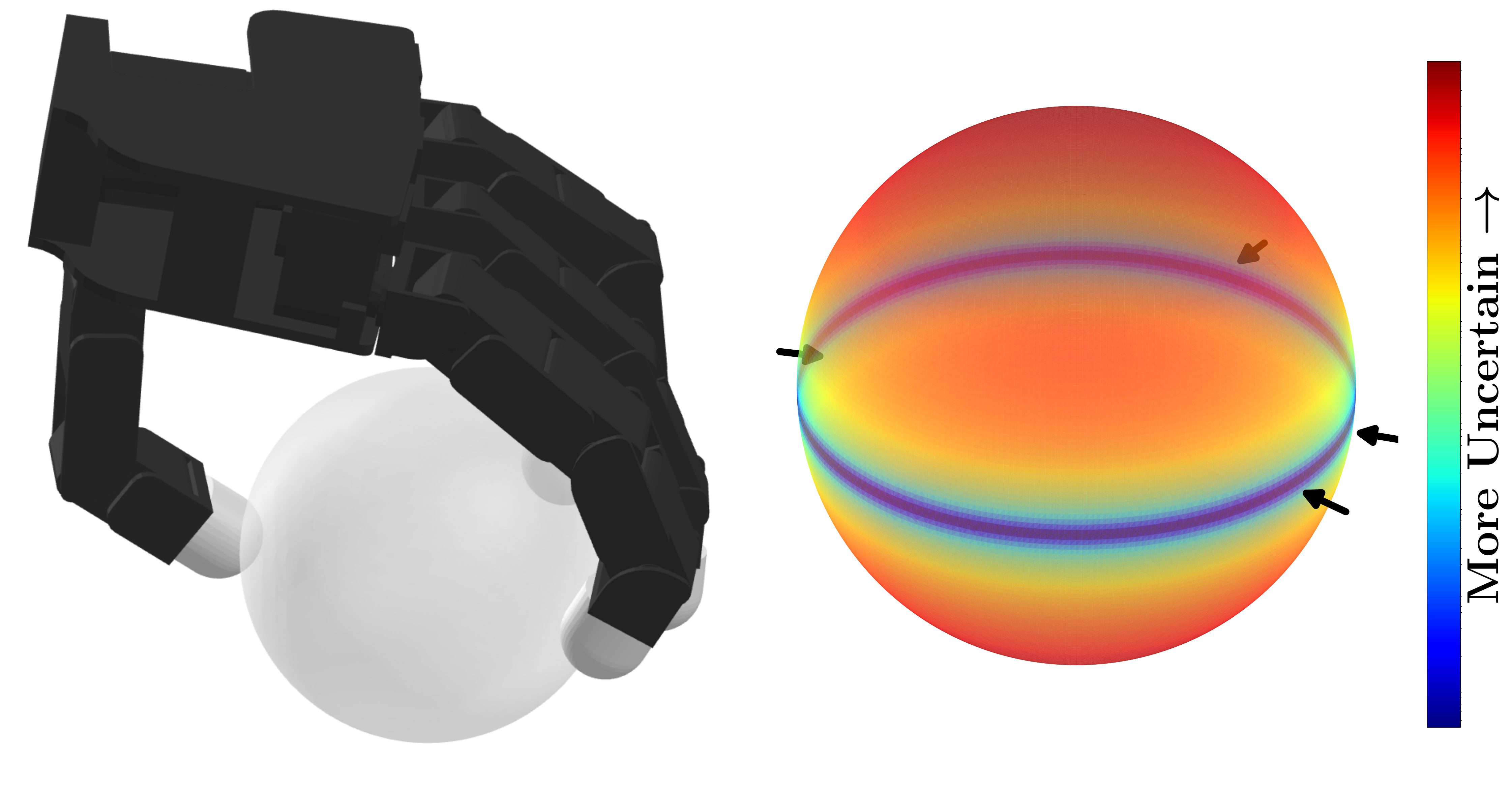}
    \caption{PONG synthesizes \textit{uncertainty-aware} dexterous precision grasps by extending the concept of \textit{force closure} to the probabilistic setting. In this example, we artificially impose uncertainty in the target object's surface normals, and PONG computes a locally-optimal grasp with fingertips near the equator, the region with the least uncertainty.}
    \label{fig:banner}
    \vspace{-0.5cm}
\end{figure}

\subsection{Related Work}
While this paper presents a novel analytic treatment of the \textit{probability of force closure} (PFC), prior work has studied PFC from data-driven and empirical perspectives. For example, \cite{weisz2012_pfc_pose_robust} notes that the traditional Ferrari-Canny epsilon-ball metric is sensitive to uncertainty in the object pose; they generate a Monte Carlo estimate of PFC via simulation, and use this to rank grasps by robustness. \textit{Gaussian process implicit surfaces} (GPISs) \cite{dragiev2011_gpis} have also been an influential way to represent the object geometry itself as uncertain, and have been deployed successfully for parallel-jaw grasping \cite{mahler2015_gpis}, dexterous grasping with tactile sensors \cite{defarias_tactile_gpis_uncertain, siddiqui2021_grasp_bo_tactile}, and for control synthesis \cite{li2016_shape_uncertainty}. While GPISs can reason jointly over both uncertain contact points and normals, they have several drawbacks, including being difficult to supervise from sensor data (requiring ground-truth signed-distance labels), poor scaling with respect to amount of data, and lack of expressivity (due to the strong smoothness prior imposed on the object surface by typical choices of the kernel function). 

An alternate approach is to instead directly learn probabilistic grasp metrics from data. For example, Dex-Net 2.0 predicts the robustness of a batch of planar parallel-jaw grasps, then selects the most robust one \cite{mahler2017_dexnet2}. A similar idea has been applied to multifinger hands in a gradient-based optimization setting with joint limits by leveraging the differentiability of neural networks \cite{lu2018_planningmultifinger, lu2020_diffgrasplearning}.

This paper is most similar in spirit to recent work on differentiable approximations of analytic metrics for fast gradient-based grasp synthesis. Proposed methods include solving sequences of SDPs \cite{dai2015_forceclosuresdp}; solving a sum of squares program \cite{liu2020_deepdiffgrasp}; optimizing a differentiable relaxation of the force closure condition \cite{liu2021_diversediffgrasps, wang2022_dexgraspnet}; solving a bilinear optimization program with a QP force closure constraint \cite{wu2022_learningdexgraspsgenmodel}; and maximizing an almost-everywhere differentiable proxy for the Ferrari-Canny metric in a bilevel setting \cite{li2023_frogger}. The common theme of these works is the use of nonlinear optimization to enforce (or penalize) kinematic, collision, and contact constraints. Our metric, PONG, is computed by solving LPs, allowing it to serve as an objective for similar bilevel programming approaches to grasp synthesis.

\subsection{Preliminaries}
We consider grasp planning for a fixed-base, fully-actuated rigid-body serial manipulator with a multi-finger hand. Denote the robot configuration $q\in\mathcal{Q}$. Assume the hand has $n_f$ fingers contacting the object at points $\{\pos^i\}_{i=1}^{n_f}$ with inward pointing surface normals $\{n^i\}_{i=1}^{n_f}$. Denote the forward kinematics maps $\pos^i=FK^i(q)$ with corresponding (translational) Jacobians $J^i(q)\in\R^{3\times n}$. Denote the single rigid object $\mathcal{O}$ with surface $\partial \mathcal{O}$.

As shorthand, let $\mathcal{C}_x:=\conv(\{x_l\}_{l})$ denote the convex hull of a finite set of points indexed by $l$. Define the \textit{wedge} operator $\wedgeop{\cdot}: \R^3 \rightarrow \mathfrak{so}(3)$ such that $\wedgeop{a} b := a \times b$ for $a,b\in\R^3$. We parameterize grasps with an optimized \textit{feasible} configuration $q^*$, which means no undesired collisions while contacting the object. We model the fingers as point contacts with Coulomb friction.

Recall that under the Coulomb friction model, a contact force $f$ satisfies the \textit{no-slip} condition if for a fixed friction coefficient $\mu$, $\norm{f_t}\leq\mu\cdot f_n$, where $t$ and $n$ denote tangent and (positive) normal components of force $f$ respectively. We call such forces \textit{Coulomb-compliant}. A force applied at point $x$ induces the torque $\tau=x \times f$, which in turn induces a corresponding wrench $w=(f, \tau)$.

The \textit{friction cone} at a point $x$ is the cone centered around the surface normal $n$ consisting of all Coulomb-compliant forces that can be applied. It is common to consider a pyramidal approximation of this cone \cite{murray1994_manipulation} with $n_s$ sides. We call its edges \textit{basis forces} and the induced wrenches \textit{basis wrenches}. We emphasize that the basis wrenches depend on the basis forces which themselves depend on the surface normal at a point $x$, a relation we use heavily below.

Let $n_w=n_cn_s$ denote the number of basis wrenches (indexed $w_j^i$), each associated with a finger $i$ and pyramid edge $j$. For clarity, we sometimes combine the indices $i,j$ into a single one $l$. Further, let $\mathcal{I}=\{1,\dots,n_f\}$, $\mathcal{J}=\{1,\dots,n_s\}$, and $\mathcal{L}=\{1,\dots,n_w\}$ denote finger, pyramid edge, and basis wrench index sets respectively.

\section{A Probabilistic Notion of Force Closure}\label{sec:pfc_bound}
This paper extends the classical theory of \textit{force closure} to the probabilistic setting in which the surface normals $n^i$ are random variables. To review, we say a grasp is (deterministically) \textit{force closure} if for any disturbance wrench it can generate Coulomb-compliant forces on the object to resist the disturbance. A well-known sufficient condition for force closure is that the origin is contained in the convex hull of a grasp's basis wrenches, i.e., $0\in\mathcal{C}_w$ \cite{rimon2019_manipulationbook}. We say that such basis wrenches \textit{certify} force closure.

In the probabilistic setting, the normals $n^i$ are random variables, so the basis forces and wrenches are as well. This begs the central question of the paper: given known contact locations and randomly-distributed surface normals, what is (a bound on) the probability that the induced basis wrenches certify force closure?

Let the known contacts be denoted $\posSet=\{\pos^1,\dots,\pos^{n_f}\}$ and consider the set of normals that induce basis wrenches certifying force closure. We call this the \textit{force closure set}:
\eqs{
    \mathfrak{N}(\posSet):=\braces{\{n^i\}_{i=1}^{n_f} \mid 0\in\mathcal{C}_w}. \label{eqn:fc}
}
Suppose the normals are jointly distributed with some density function $\pdf\parens{n^1, \ldots, n^{n_f}}$ and that they are mutually independent such that $\pdf$ can be factorized as $\prod_{i=1}^{n_f} \pdf(n^{i})$. Then, the probability of force closure can be computed by the integral
\eqs{
    P_\text{fc} = \mathbb{P}[0\in\mathcal{C}_w] = \int_{\mathfrak{N}(\posSet)} \prod_{i=1}^{n_f} \pdf(n^i) dn^i. \label{eqn:pfc_intractable}
}

While succinct, \eqref{eqn:pfc_intractable} is not directly useful for two reasons. First, integrating the density $p$ over $\mathfrak{N}$ is difficult, because even though $\pdf$ is factorizable, the integral itself is not. The integration variables are coupled by the domain $\mathfrak{N}$, since $\mathcal{C}_w$ depends on all of the random normals. Second, $\mathfrak{N}$ is difficult to parameterize; it has no closed form since it is implicitly defined by the convex hull condition \eqref{eqn:fc}, where hull membership is checked by solving an LP \cite{rimon2019_manipulationbook}.

Thus, our strategy is to derive an \textit{approximate force closure set} $\mathcal{A}\subseteq\mathfrak{N}$ in tandem with a choice of density function $p$ for which the integral over $\mathcal{A}$ is known. Since $\mathcal{A}\subseteq\mathfrak{N}$, integrating $\pdf$ over $\mathcal{A}$ yields a lower bound on $P_\text{fc}$.

The following result \cite{hayashi2017_bivariategaussianintegral} will motivate our constructions by providing an exact method for integrating a bivariate Gaussian density over an arbitrary planar polygon.

\begin{figure*}
    \centering
    \includegraphics[width=\linewidth]{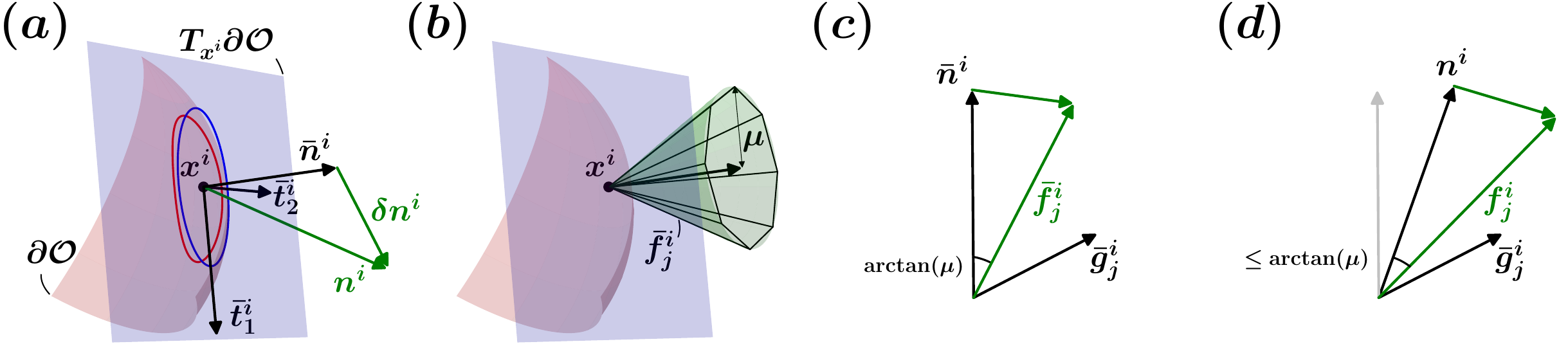}
    \caption{\textbf{(a)} The parameterization from Sec. \ref{subsec:uncertainty_parameterization}. For visual clarity, we point the contact normals outward. The variances defining the uncertainty ellipse (blue) lie in the tangent plane at point $x^i$ on surface $\partial\mathcal{O}$. A perturbation $\delta n^i$ is drawn to generate a random normal $n^i$. \textbf{(b)} The friction cone (green) and a canonical pyramidal approximation. Each basis wrench $\bar{w}_j^i$ depends on position $x^i$, surface normal $\bar{n}^i$, and friction coefficient $\mu$. \textbf{(c)} The construction of a mean basis force $\bar{f}^i_j$ using generator $\bar{g}^i_j$. \textbf{(d)} The construction of a random basis force $\bar{f}^i_j$ using the same generator as in the mean case. Note the smaller angle, which ensures the random basis forces are Coulomb-compliant.}
    \label{fig:pyramid}
    \vspace{-0.5cm}
\end{figure*}

\begin{proposition}\label{prop:bigauss}
    Let $\mathscr{P}$ be a planar polygon with $M$ vertices $y^1,\dots,y^M\in\R^2$ ordered counterclockwise with $y^{M+1}:=y^1$. Let $Z$ be a bivariate Gaussian random variable with mean $\mu$ and covariance $\Sigma=\diag(\sigma_1^2,\sigma_2^2)$. Then, 
    \eqs{\label{eqn:integral}
        \mathbb{P}[Z\in\mathscr{P}] = \frac{1}{\sigma_2\sqrt{8\pi}}\sum_{m=1}^M D^m\int_0^1A^m(r)B^m(r)dr,
    }
    where the terms above are
    \eqsnn{
        D^m &:= y_2^{m+1} - y_2^m, \\
        A^m(r) &:= \exp\parens{-\frac{1}{2\sigma_2^2}\brackets{(1-r)y_2^m + ry_2^{m+1} - \mu_2}^2}, \\
        B^m(r) &:= \erf\parens{\frac{(1-r)y_1^m+ry_1^{m+1} - \mu_1}{\sigma_1\sqrt{2}}}.
    }
\end{proposition}

This allows us to numerically integrate over planar regions with just line integrals (full proof in \ref{app:bigauss_proof}).

\section{PONG: A Tractable PFC Lower Bound} \label{sec:pong}

This section presents PONG, which leverages Proposition \ref{prop:bigauss} to address the aforementioned challenges. Due to space constraints, we elide computational details to our open-source implementation and the Appendix.

We proceed in three steps: (A) we present a construction that linearly relates a random surface normal $n^i$ to a set of random basis wrenches $w^i_j$; (B) we use this relation to define disjoint sets $\mathcal{A}_i$ such that $\mathcal{A}=\bigcup_i\mathcal{A}_i$, which allows us to factorize the integral of $\pdf$ over $\mathcal{A}$; and (C) we parameterize the sets $\mathcal{A}_i$ in an integrable way and compute them with linear programming. The end result is the lower bound
\eqs{\label{eqn:lfc}
    L_\text{fc} := \prod_{i=1}^{n_f}\int_{\mathcal{A}^i}p(n^i)dn^i &= \int_{\mathcal{A}}\prod_{i=1}^{n_f}p(n^i)dn^i \\
    &\leq \int_{\mathfrak{N}}\prod_{i=1}^{n_f}p(n^i)dn^i,
}
which is the main result of the paper.

\subsection{Random Normals, Forces, and Wrenches}\label{subsec:uncertainty_parameterization}

We begin with our representation of the random surface normals $n^i$. We model each random normal as the sum of a deterministic \textit{mean normal vector} denoted $\bar{n}^i$ and a \textit{random perturbation vector} $\delta n^i = \mathcal{T}_i \epsilon,$ where $\mathcal{T}_i = \begin{bmatrix}\bar{t}^i_1 & \bar{t}^i_2\end{bmatrix} \in \mathbb{R}^{3 \times 2}$ is a basis for the tangent plane at $\bar{n}^i$, and $\epsilon \sim \mathcal{N}(0, \Sigma^i)$ is a zero-mean Gaussian random vector in $\mathbb{R}^2$ (see Fig. \ref{fig:pyramid}a).

We opt to parameterize the uncertain normals in this way for two reasons. First, all perturbations in the direction of $\bar{n}^i$ do not change the direction of the mean normal, and therefore do not represent any uncertainty in its orientation. Second, the planar restriction will allow us to invoke the tractable Gaussian density integral given in Proposition \ref{prop:bigauss}.

We remark that with this construction, the random normals $n^i$ will not have unit norm. However, since friction cones are invariant under scaling, as long as the random basis forces $f_j^i$ remain Coulomb-compliant, we can still certify force closure with the induced random basis wrenches.

Leveraging this insight, we now introduce a procedure for constructing a random friction pyramid about the random normal $n^i$ such that its edges are always Coulomb-compliant. First, consider the \textit{mean} basis wrenches $\bar{w}_j^i$ with force and torque components $\bar{f}_j^i$ and $\bar{\tau}_j^i$. Per Fig. \ref{fig:pyramid}b, we represent $\bar{f}_j^i$ as the sum of $\bar{n}^i$ and a tangent component $\parens{f_j^i}_t$ of length $\mu$. We compute $\parens{f_j^i}_t$ using a unit length \textit{generator} $\bar{g}_j^i(\bar{n}^i)\in\R^3$ (see Fig. \ref{fig:pyramid}c) of our choice orthogonal to $\bar{n}^i$ such that
\eqs{
    \parens{f_j^i}_t = \mu\parens{\bar{g}_j^i \times \bar{n}^i}.
}
For example, for some arbitrary $v\neq\bar{n}^i$, we could pick $\bar{g}_j^i(\bar{n}^i)=(v \times \bar{n}^i)/\norm{v \times \bar{n}^i}_2$. Thus, we can write
\eqs{
    \bar{f}_j^i &= \bar{n}^i + \mu\parens{\bar{g}_j^i \times \bar{n}^i} \\
    \implies \bar{w}_j^i &= \underbrace{\mat{\parens{I + \mu\wedgeop{\bar{g}_j^i}} \\ \wedgeop{x^i}\parens{I + \mu\wedgeop{\bar{g}_j^i}}}}_{:=T_j^i(\bar{n}^i)}\bar{n}^i.
}

Generally, $\bar{w}_j^i$ is nonlinear with respect to $\bar{n}^i$ since the generators can depend nonlinearly on $\bar{n}^i$. However, if we choose to generate the \textit{random} basis wrenches $w^i_j$ with the generators from the mean case, we have the linear relation $w_j^i = T_j^i(\bar{n}^i)n^i$ with respect to random normal $n^i$.

Our construction satisfies $\norm{(f_j^i)_t}_2\leq\mu\norm{n^i}$ $\big($in contrast, observe that $\norm{\bar{f}_j^i}_2=\mu\norm{\bar{n}^i}$ as in Fig. \ref{fig:pyramid}c and \ref{fig:pyramid}d$\big)$. To see this, let $\psi_j^i$ be the angle between $\bar{g}_j^i$ and $n^i$. Then,
\eqs{
    \norm{(f_j^i)_t}_2 = \mu\norm{\bar{g}_j^i}\norm{n^i}\sin(\psi_j^i)\leq\mu\norm{n^i},
}
ensuring Coulomb-compliance of the random basis forces.

\subsection{Deriving a Decomposable Approximate Force Closure Set}\label{subsec:decoupling}
Next, we provide a procedure to generate disjoint sets $\mathcal{A}_i$ whose union defines the inner approximation $\mathcal{A}$ of the ideal integration domain $\mathfrak{N}$, allowing us to separate the $P_\text{fc}$ integral \eqref{eqn:pfc_intractable}. We will use the following result, which certifies the random basis wrenches $w^i_j$ form a force closure grasp.

\begin{figure}
    \centering
    \includegraphics[width=\linewidth]{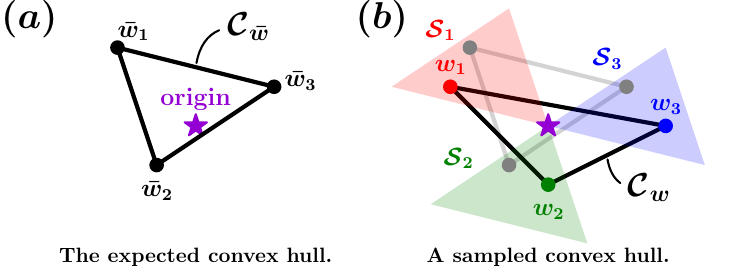}
    \caption{Prop. \ref{prop:containment} can be interpreted with the sets $\mathcal{S}_l:=-\mathcal{C}_{\bar{w}}+\bar{w}_l$ (colored) centered about the expected wrenches. If the random wrenches satisfy $w_l\in\mathcal{S}_l,\;\forall l\in\mathcal{L}$, then $0\in\mathcal{C}_w$, so the sampled wrenches certify force closure.
    }
    \label{fig:lemma}
    \vspace{-0.5cm}
\end{figure}

\begin{proposition}\label{prop:containment}
    If $w_l-\bar{w}_l\in -\mathcal{C}_{\bar{w}}$ for all $l\in\mathcal{L}$, then $0\in\mathcal{C}_w$.
\end{proposition}
\begin{proof}
    Suppose for the sake of contradiction that $0\not\in\mathcal{C}_w$. Then, there exists $a$ such that $\inner{a}{w}>0$ for all $w\in\mathcal{C}_w$ by the separating hyperplane theorem. Further, there must exist some $l^*$ satisfying $\inner{a}{\bar{w}_{l^*}}\leq\inner{a}{\bar{w}_{l}}$ for all $\bar{w}_l\in\mathcal{C}_{\bar{w}}$. Since $w_l-\bar{w}_l\in-\mathcal{C}_{\bar{w}}$ for each $l$, we have corresponding to $l^*$ a set of convex weights $\alpha^{l^*}\in\R^{n_w}$ such that
    \eqs{\label{eqn:lemma_eqn}
        w_{l^*} = \bar{w}_{l^*} &+ (w_{l^*} - \bar{w}_{l^*}) = \bar{w}_{l^*}-\sum_{l=1}^{n_w}\alpha^{l^*}_l\bar{w}_l \\
        \implies \inner{a}{w_{l^*}} &= \inner{a}{\bar{w}_{l^*}} - \sum_{l=1}^{n_w}\alpha^{l^*}_l\inner{a}{\bar{w}_l} \\
        &\leq \inner{a}{\bar{w}_{l^*}} - \inner{a}{\bar{w}_{l^*}}\sum_{l=1}^{n_w}\alpha^{l^*}_l = 0.
    }
    But, $\inner{a}{w_{l^*}}>0$ since $w_{l^*}\in\mathcal{C}_w$, contradicting \eqref{eqn:lemma_eqn}.
\end{proof}
We view $w_l-\bar{w}_l$ as the deviation of the $l^{th}$ basis wrench from some mean value (e.g., from a model). If all deviations are contained in $-\mathcal{C}_{\bar{w}}$, then by Proposition \ref{prop:containment}, the origin lies in the convex hull $\mathcal{C}_w$ of the true, random basis wrenches, i.e., \textit{the grasp is force closure}. See Fig. \ref{fig:lemma} for visual intuition.

Combining this with the relation $w^i_j = T^i_jn^i$ from Sec. \ref{subsec:uncertainty_parameterization}, we thus want our approximations to satisfy
\begin{equation}\label{eqn:decomposed_fc_sets}
    \mathcal{A}_i \subseteq \{ n^i \mid T^i_j (n^i - \bar{n}^i) \in \mathcal-{C}_{\bar{w}},\;\forall j\in\mathcal{J}\}.
\end{equation}
By applying Proposition \ref{prop:containment}, we have
\eqs{
    n^i\in\mathcal{A}^i,\;\forall i\in\mathcal{I} \implies (n^1,\ldots,n^{n_f})\in\mathfrak{N},
}
so letting $\mathcal{A}=\bigcup_{i=1}^{n_f}\mathcal{A}_i$, we recover the bound \eqref{eqn:lfc}. The decomposition is possible because even though in \eqref{eqn:decomposed_fc_sets}, $\mathcal{C}_{\bar{w}}$ depends on all the \textit{mean} surface normals $(\bar{n}^1,\ldots,\bar{n}^{n_f})$, these are fixed parameters of the known uncertainty distributions. In contrast, in \eqref{eqn:fc}, $\mathcal{C}_w$ (and thus $\mathfrak{N}$) depends jointly on all of the \textit{random} normals, which are the integration variables.

\subsection{Computing Approximate Force Closure Sets}
Unfortunately, the condition on the right hand side of \eqref{eqn:decomposed_fc_sets} cannot be expressed in closed form. Thus, we let $\mathcal{A}_i$ be a conservative polygonal approximation of it by performing a search procedure to find a polytope with large volume satisfying \eqref{eqn:decomposed_fc_sets} parameterized by its vertices.

To do this, for each finger, we fix a set of search directions $d^{i,k}\in\R^3$ in the tangent plane at $x^i$. We then search for the longest step $\theta^{i,k}\geq0$ that can be taken in this direction while satisfying \eqref{eqn:decomposed_fc_sets}. We denote the point $\theta^{i,k} d^{i,k}$ as $v^{i,k}$ and let $\mathcal{A}_i$ be the convex hull of these points. Since the set on the right side of \eqref{eqn:decomposed_fc_sets} is convex, $\mathcal{A}_i$ is a conservative approximation.

Conveniently, this search can be expressed as a linear program. Define the matrix $\widebar{W}\in\R^{6\times n_w}$ whose columns are the mean basis wrenches. Then, to compute each $v^{i,k}$, we can solve the following LP, denoted VLP$^{i,k}$:
\begin{subequations}
\label{opt:vlp}
\begin{align}
    \maximize_{\substack{\theta^{i,k} \in \R \\ \braces{\alpha^{i,k}_j}_{j=1}^{n_s} \subset \R^{n_w}}} \quad & \theta^{i,k} \\
    \subto \quad & \theta^{i,k} \geq 0 \label{constr:vlp_delta} \\
    & \alpha^{i,k}_j \succeq 0,\;\forall j\in\mathcal{J} \label{constr:vlp_alpha1} \\
    & \mathds{1}^{\top}_{n_w} \alpha^{i,k}_j = 1,\;\forall j\in\mathcal{J} \label{constr:vlp_alpha2} \\
    & T_j^i\parens{\theta^{i,k} d^{i,k}} = -\widebar{W}\alpha^{i,k}_j,\;\forall j\in\mathcal{J}. \label{constr:cvx}
\end{align}
\end{subequations}
Constraint \eqref{constr:vlp_delta} enforces nonnegativity of the scaling along $d^{i,k}$, constraints \eqref{constr:vlp_alpha1} and \eqref{constr:vlp_alpha2} enforce that the $\alpha^{i,k}_j$ are valid convex weights, and constraint \eqref{constr:cvx} enforces that the random normal $n^i = \bar{n}^i + \theta^{i,k} d^{i,k}$ must lie in $\mathcal{A}_i$. 

\begin{figure}[t]
    \centering
    \includegraphics[width=\linewidth]{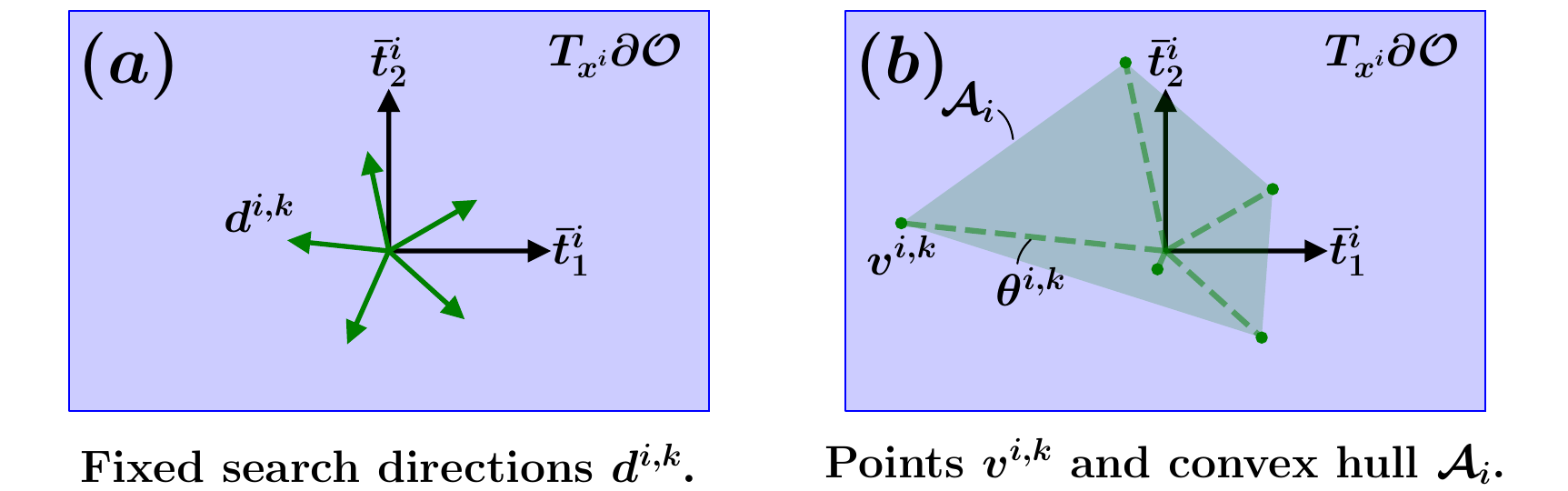}
    \caption{Visualization of the parameterization and construction of $\mathcal{A}_i$. \textbf{(a)} Example search directions represented in the planar coordinates of the tangent basis. \textbf{(b)} Example points placed along the search directions and the corresponding set $\mathcal{A}_i$.
    }
    \label{fig:vertices}
    \vspace{-0.5cm}
\end{figure}

\begin{proposition}\label{prop:vlp_feasibility}
    VLP$^{i,k}$ has a feasible solution if $0\in\mathcal{C}_{\bar{w}}$.
\end{proposition}
\begin{proof}
    If $0\in\mathcal{C}_{\bar{w}}$, then $\exists\tilde{\alpha}\in\R^{n_w}$ such that $\tilde{\alpha}\succeq0, \bar{W}\tilde{\alpha}=0,$ and $\mathds{1}^{\top}_{n_w}\tilde{\alpha}=1$. Thus, the choices $\delta^{i,k}=0$ and $\alpha^{i,k}_j=\tilde{\alpha},\;\forall j$ satisfy \eqref{constr:vlp_delta}-\eqref{constr:cvx}, so \eqref{opt:vlp} is feasible.
\end{proof}

Proposition \ref{prop:vlp_feasibility} ensures that if a grasp is force closure in the mean case, then we can feasibly solve a batch of VLPs in parallel over all $(i,k)\in\mathcal{I}\times\mathcal{K}$ and then recover a nontrivial value for $L_\text{fc}$ using the expressions in Proposition \ref{prop:bigauss}. If any VLPs in the batch are infeasible, we set the bound to 0 and do not compute any integrals.

\section{Applying PONG to Synthesize Curvature-Regularized Grasps} \label{sec:application}
A common failure mode for grasp synthesis is \textit{edge-seeking}, in which a grasp optimizer computes a solution with the fingertips on sharp edges of the object.
We hypothesize two causes. First, in the case of, e.g., the Ferrari-Canny metric, when the torque origin lies near the object's center, the edges yield larger moment arms, theoretically providing more robustness. Second, if an initial guess for a grasp is far from the object, the edges are often the closest points to the fingertips, and edge points are often the first feasible points encountered during optimization; thus, they easily become local minima for the full optimization.

Edges (and more generally high-curvature areas) tend to be poor grasp locations, because small errors in perception lead to failure. In this section, we use PONG to generate \textit{curvature-regularized} grasps by artificially assigning high uncertainty to high-curvature areas.

\begin{figure}
    \centering
    \includegraphics[width=\linewidth]{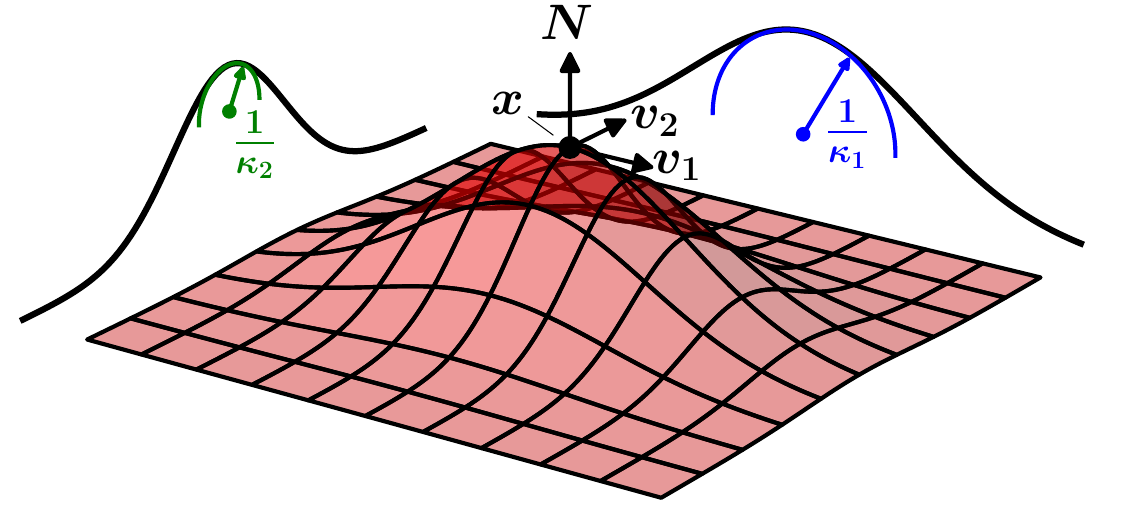}
    \caption{A visualization of the shape operator. Shown is the surface normal $N$, the principal directions $v_1, v_2$, and the principal curvatures $\kappa_1,\kappa_2$. Note the close resemblance to our uncertainty parameterization in Fig. \ref{fig:pyramid}a.}
    \label{fig:curvature}
    \vspace{-0.5cm}
\end{figure}

\subsection{Curvature-Based Uncertainty Distributions}
Suppose we represent the surface $\partial\mathcal{O}$ as the 0-level set of a smooth function $s:\R^3 \rightarrow \R$. At a point $x\in\partial\mathcal{O}$, the \textit{shape operator} $\mathfrak{S}_x:T_x\partial\mathcal{O} \rightarrow T_x\partial\mathcal{O}$ is defined $\mathfrak{S}_x(v):=-\nabla_vN(x)$, where $N$ is the unit surface normal at $x$. The two eigenpairs of $\mathfrak{S}_x$, $(\kappa_1,v_1),(\kappa_2,v_2)\in \R \times T_x\partial\mathcal{O}$, are the \textit{principal curvatures and directions} at $x$ respectively \cite{oneill2006_diffgeom}.

Letting the principal directions $v_1,v_2$ form our tangent basis at each contact $x^i$ and (some function of) the magnitude of the principal curvatures $\kappa_1,\kappa_2$ form our variances, we can recover a curvature-sensitive uncertainty distribution.

Since $\mathfrak{S}_x(v)=-\brackets{\nabla^2s(x)}(v)$, the eigenpairs of $\mathfrak{S}_x$ are those of the Hessian of $s$. In the sequel, we use the following uncertainty distribution parameters:
\eqs{\label{eqn:curvature_distribution}
    \bar{n}^i &:= -\nabla s(x^i), \\
    \bar{t}^i_m &:= v_m,\; m\in\{1,2\}, \\
    \parens{\sigma^i_m}^2 &:= \log\parens{K_{\text{curv}}\cdot\abs{\kappa_{m}} + \epsilon},\;m\in\{1,2\},
}
where $K_{\text{curv}}>0$ is a parameter relating curvature to uncertainty and $\epsilon>0$ captures prior uncertainty at $x$. These values are also defined for points $x\not\in\partial\mathcal{O}$, i.e., the distribution is even defined ``off-surface,'' which we exploit during optimization to evaluate infeasible iterates.

\subsection{Grasp Synthesis with Nonlinear Optimization}
As a test, we synthesized grasps by solving the following bilinear optimization program (similar to FRoGGeR \cite{li2023_frogger}):
\begin{subequations}\label{opt:curvature_pong}
\begin{align}
    \maximize_{q\in\mathcal{Q}}\quad& L_\text{fc}(q) \label{opt:curvature_pong_a} \\
    \subto \quad & q_{min} \preceq q \preceq q_{max} \label{opt:curvature_pong_b} \\
    &\bar{\ell}^*(q) \geq 0.3 \label{opt:curvature_pong_c} \\
    & s(FK^i(q))=0,\; i=1,\dots,n_c \label{opt:curvature_pong_d} \\
    & \sigma\parens{o_A^{(m)}, o_B^{(m)};q} \geq d_m,\; m=1,\dots,n_p. \label{opt:curvature_pong_e}
\end{align}
\end{subequations}
In \eqref{opt:curvature_pong_c}, $\bar{\ell}^*(q)$ is the \textit{normalized min-weight metric} \cite{li2023_frogger}. If $\bar{\ell}^*(q)>0$, then $0\in\mathcal{C}_{\bar{w}}$, so whenever \eqref{opt:curvature_pong_c} is satisfied, we can invoke Proposition \ref{prop:vlp_feasibility} to feasibly solve all VLPs.

Constraint \eqref{opt:curvature_pong_d} enforces that the contacts $x^i$ lie on the surface, where $s$ is a neural implicit surface trained using volume rendering \cite{wang2021_neus}. Lastly, $\sigma$ is a signed distance between two convex collision geometries, $d_m$ is the minimum allowable distance between geometries in pair $m$, and $n_p$ is the total number of collision pairs (see \cite{li2023_frogger} for details).

These pairs are computed by performing a convex decomposition of the object using VHACD \cite{mamou2009_vhacd}. Unlike \cite{li2023_frogger}, we \textit{do not} prescribe contact points $x^i$, so when the fingers are not on the surface, we let $x^i$ be the closest point on the fingertip geometry to the object, computed using \verb|Pinocchio| \cite{carpentier2019_pinocchio}.

\begin{figure}
    \centering
    \includegraphics[width=\linewidth]{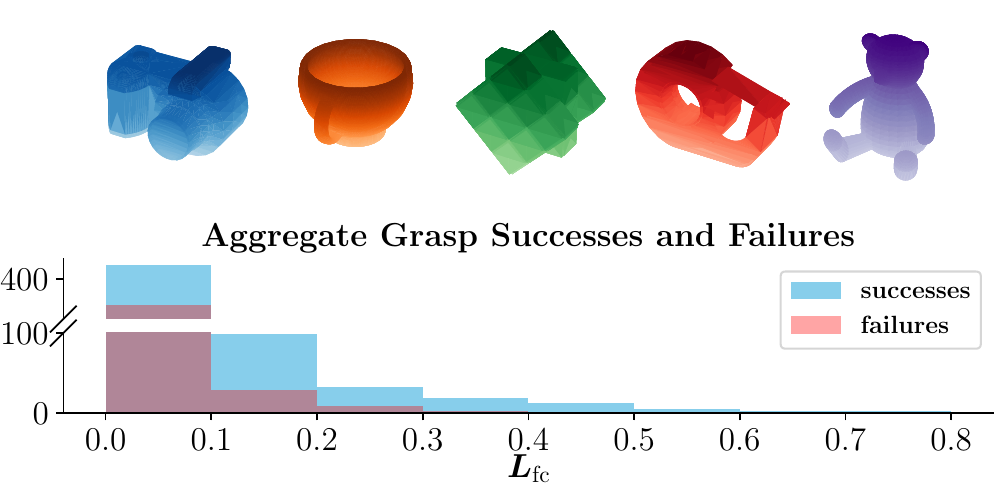}
    \caption{Simulation results. \textbf{Top.} The 5 objects used for simulations (camera, teacup, Rubik's cube, tape dispenser, teddy bear). They were chosen due to their high nonconvexity and many edges. \textbf{Bottom.} Success/failure histograms over all trials. Note the fast decay of failures relative to successes.}
    \label{fig:sim}
    \vspace{-0.5cm}
\end{figure}

\begin{figure*}
    \centering
    \includegraphics[width=\linewidth]{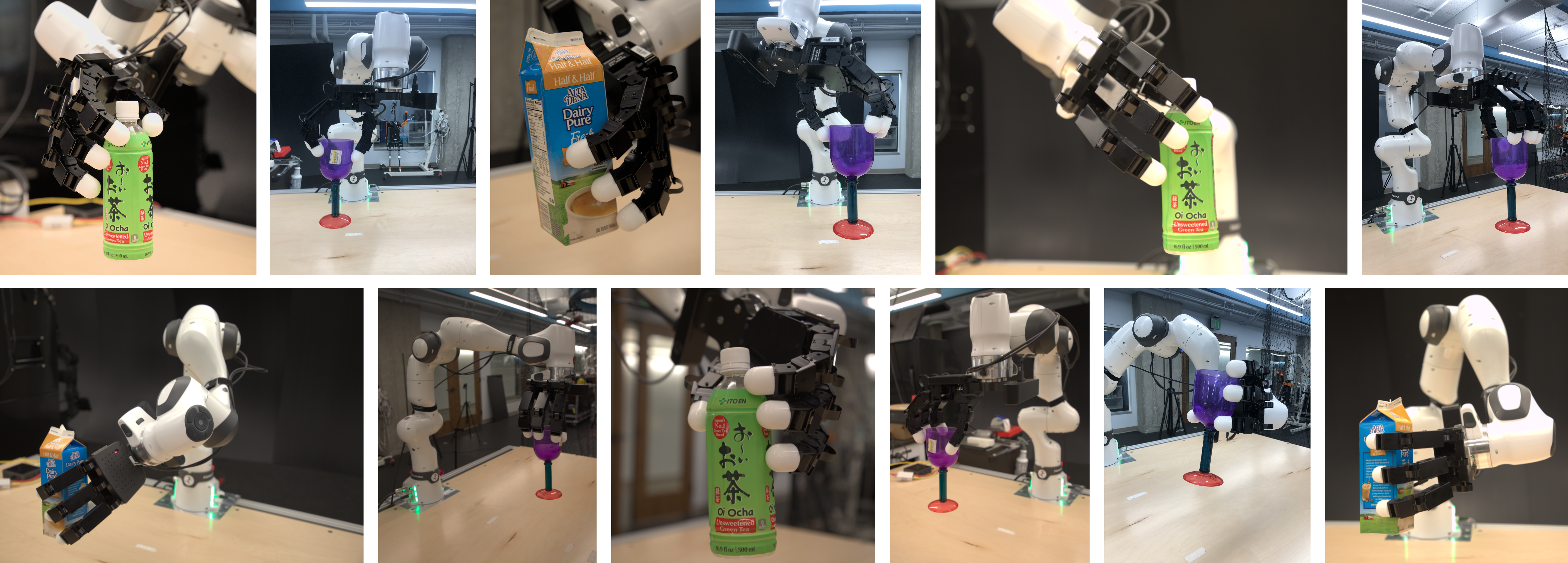}
    \caption{We used an Allegro hand with an attached Zed camera to image each object and deployed PONG to successfully synthesize diverse precision grasps even in the presence of transparency, reflectivity, and an abundance of edges. Shown are representative successful picks.}
    \label{fig:hardware}
    \vspace{-0.5cm}
\end{figure*}

\subsection{Simulation Results}\label{sec:sim}
Based on the theory of Sec. \ref{sec:pfc_bound}, we expect fewer grasp failures as $L_\text{fc}$ rises. To test this, we synthesized grasps on each of 5 nontrivial objects (see Fig. \ref{fig:sim}) and visualized the distribution of successes and failures on a ``shaky pickup'' task, where we added high-frequency sinusoidal perturbations to a straight reference trajectory with an amplitude of 3\verb|cm|. We simulated 100 unperturbed and 100 perturbed grasps per object for a total of 1000 grasps.

We use the following simple controller: for an optimal solution $q^*$ of \eqref{opt:curvature_pong}, we have corresponding fingertip positions $x^i$ and mean normals $\bar{n}^i$ computed using \eqref{eqn:curvature_distribution}. We define new target positions for each finger 1\verb|cm| into the surface,
\eqs{
    x_{\text{new}}^i = x^i + 0.01\nabla s(x^i),
}
and let $q^*_{\text{new}}$ satisfy $x_{\text{new}}^i = FK^i(q^*_{\text{new}}),\forall i\in\mathcal{I}$, computed using inverse kinematics. Lastly, we track $q^*_{\text{new}}$ with P control:
\eqs{
    \tau_{\text{fb}} = -K_p(q - q^*_{\text{new}}).
}

Shown in Fig. \ref{fig:sim} are the aggregate successes and failures over all 1000 grasps. We highlight two observations: (1) as $L_\text{fc}$ increases, the failure rate decays as predicted; and (2) the bound's conservativeness is nontrivial, as even in the regime where $L_\text{fc}\approx0$, about half of the grasps are successful. That is, if $L_\text{fc}$ is high, we should be confident in a grasp's quality, but the converse does not hold.

We report an aggregate success rate of 59\%, a low mark that increases to 75\% (266/354) if we exclude very low-confidence grasps ($L_\text{fc} < 0.05$). We find that after these exclusions, the remaining grasps are resistant to perturbations, with a success rate of 76.8\% (136/177) when unperturbed and 73.4\% (130/177) when perturbed, even though aggregately, the perturbations induced a 6.4\% drop in success rate. This suggests that many of the induced failures occur in the low-confidence regime, which supports the hypothesis that curvature regularization improves grasp robustness.

Finally, over all trials, the median grasp synthesis time was 5.93\verb|s|, a result of our optimized implementation of PONG, particularly via parallel LP solves.



\subsection{Hardware Results}
\label{subsec:hardware}
We additionally verified that we could use PONG to generate grasps on a real robotic system by synthesizing 10 grasps for each of 3 objects: a semi-transparent tea bottle, an opened (empty) milk carton, and a transparent/reflective colored goblet. To achieve this, we chose to represent each object as a \textit{neural radiance field} (NeRF) \cite{mildenhall2020_nerf}, from which we extracted a smooth density field defined for any query point $x\in\R^3$. We then represented the surface $\partial\mathcal{O}$ as a level set of this density function, where the chosen level was selected experimentally for each object. The images used to train each NeRF were collected by a wrist-mounted camera which captured images of the object by following a fixed trajectory. We then used this noisy representation to solve problem \eqref{opt:curvature_pong}, yielding an optimal grasp configuration $q^*$.

Finally, we planned a pick trajectory to achieve the optimized grasp. We report pick success rates of 6/10, 7/10, and 7/10 respectively, which we qualify with the following remarks. First, the NeRF representations exhibited significant noise, often with many portions of the object surface missing due to limited views of each object. Second, we deployed the naive ``open-loop'' grasp controller from Sec. \ref{sec:sim}, which usually fails in the presence of large perception errors. Finally, pick success was very sensitive to the choice of initial condition for the optimizer. For example, all but one failure on the tea bottle were attempted overhead grasps which tried to pinch the poorly-reconstructed cap, while all successes were side grasps. When the initial guess was similar to a good grasp, our experiments support the hypothesis that PONG will refine it into a performant, feasible one.

\section{Conclusion}\label{sec:conclusion}
In this paper, we considered the problem of grasp synthesis under surface normal uncertainty. We developed PONG, a novel, analytic lower bound on the probability of force closure (PFC), which provides a principled measure of a grasp's robustness, is fast to compute, and is differentiable. Thus, it is able to serve as an objective function for a gradient-based nonlinear optimizer. We proved that our metric is a lower bound on PFC for Gaussian-distributed surface normals. In particular, we applied PONG to the case of curvature-regularized grasping, where we used the Gaussian curvature of the object surface as an uncertainty metric, and showed that this practical choice of uncertainty distribution yields PFC bounds that are strongly correlated with grasp success/failure. Finally, we provide a hardware study of our method, using this curvature metric to optimize risk-sensitive grasps for objects represented as NeRFs. 

This work provides numerous directions for future work. One immediate direction is to consider the effect of uncertain contact locations $x^i$, which is very common with poor sensors. Another is improving the conservative nature of our bound; we make several linearizing approximations that make the bound fast to compute at the cost of weakening it.
It would be also interesting to explore data-driven methods for predicting surface normal distributions from, e.g., visual data.
Finally, as discussed in Section \ref{subsec:hardware}, a key limitation of our hardware experiments was the sensitivity of the grasps to the grasp controller itself, which does not reason about the surface uncertainty; we plan to explore tools from robust optimization, combined with the surface normal uncertainty representations developed here, to perform robust grasp force optimization for risk-sensitive grasp control or in-hand manipulation.

\newpage
\balance

\bibliographystyle{unsrt}
\bibliography{references}

\newpage

\appendix
\subsection{Proof of Proposition \ref{prop:bigauss}}\label{app:bigauss_proof}
The proof of Lemma \ref{prop:bigauss} closely follows the one presented in \cite[Proposition 1]{hayashi2017_bivariategaussianintegral}. First, we recall Green's Theorem.
\begin{theorem}[Green's Theorem]\label{thm:greens_thm}
    Let $\mathcal{D}$ be a closed region in the plane with piecewise smooth boundary. Let $P(y_1,y_2)$ and $Q(y_1,y_2)$ be continuously differentiable functions defined on an open set containing $\mathcal{D}$. Then,
    \eqs{
        &\oint_{\partial\mathcal{D}}P(y_1,y_2)dy_1 + Q(y_1,y_2)dy_2 \\
        &\quad = \iint_{\mathcal{D}}\parens{\frac{\partial Q}{\partial y_1} - \frac{\partial P}{\partial y_2}}dy_1dy_2.
    }
\end{theorem}

Second, we prove a useful intermediate result.
\begin{lemma}\label{lemma:exp_integral}
The following holds:
    \eqs{\label{eqn:lemma_integral_eqn}
        &\int\exp(-(ay^2+2by+c))dy \\
        &\quad = \frac{1}{2}\sqrt{\frac{\pi}{a}}\exp\parens{\frac{b^2-ac}{a}}\erf\parens{\sqrt{a}y + \frac{b}{\sqrt{a}}} + \textrm{const.}
    }
\end{lemma}
\begin{proof}
    Recall that
    \eqs{
        \erf(y) = \frac{2}{\sqrt{\pi}}\int_0^y\exp(-t^2)dt.
    }
    We differentiate the RHS of \eqref{eqn:lemma_integral_eqn} with respect to $y$ and by the Fundamental Theorem of Calculus, we have
    \eqs{
        &\frac{1}{\sqrt{a}}\exp\parens{\frac{b^2-ac}{a}}\cdot\sqrt{a}\exp\parens{-\parens{\sqrt{a}y + \frac{b}{\sqrt{a}}}^2} \\
        &\quad= \exp\parens{\frac{b^2-ac}{a}}\exp\parens{-\parens{ay^2+2by+\frac{b^2}{a}}} \\
        &\quad= \exp(-(ay^2+2by+c)),
    }
    which is the integrand of the LHS, proving the claim.
\end{proof}
We are now ready to prove Proposition \ref{prop:bigauss}.
\begin{proof}[Proof of Proposition \ref{prop:bigauss}]
    We have that
    \eqs{
        &(y-\mu)^\top\Sigma^{-1}(y-\mu) \\
        &\quad= \frac{1}{\sigma_1^2}y_1^2 - \frac{2}{\sigma_1^2}\mu_1y_1 + \brackets{\frac{1}{\sigma_2^2}(y_2-\mu_2)^2 + \frac{\mu_1^2}{\sigma_1^2}}.
    }
    Letting
    \eqs{
        a &= \frac{1}{2\sigma_1^2},\\
        b &= \frac{-\mu_1}{2\sigma_1^2},\\
        c &= \frac{1}{2}\brackets{\frac{1}{\sigma_2^2}(y_2-\mu_2)^2 + \frac{\mu_1^2}{\sigma_1^2}},
    }
    and applying Lemma \ref{lemma:exp_integral} to the bivariate Gaussian density function $f(y_1,y_2)$, we have
    \eqs{
        &\int f(y_1,y_2)dy_1 \\
        &= \frac{1}{2\sigma_2\sqrt{2\pi}}\exp\parens{-\frac{1}{2}\parens{\frac{y_2 - \mu_2}{\sigma_2}}^2}\erf\parens{\frac{y_1-\mu_1}{\sigma_1\sqrt{2}}} \\
        &\quad+ \textrm{const}.
    }
    Applying Theorem \ref{thm:greens_thm}, we see that for the choice $P=0$ and $Q=\int f(y_1,y_2)dy_1$,
    \eqs{
        &\iint_\mathcal{D}f(y_1,y_2)dy_1dy_2 \\
        &= \oint_{\partial\mathcal{D}}\frac{1}{2\sigma_2\sqrt{2\pi}}\exp\parens{-\frac{\parens{y_2 - \mu_2}^2}{2\sigma_2^2}}\erf\parens{\frac{y_1-\mu_1}{\sigma_1\sqrt{2}}}dy_2.
    }
    To evaluate the contour integral, we split the contour up into the line segments formed by connecting the $M$ extreme points of $\mathcal{D}$ in counterclockwise order. A point $y$ on the $m^{th}$ segment can be expressed
    \eqs{
        y = (1-r)\mat{y_1^m \\ y_2^m} + r\mat{y_1^{m+1} \\ y_2^{m+1}},\; r\in[0,1].
    }
    Performing this change of variables and letting $y^{N+1}=y^1$,
    \eqs{
        &\iint_\mathcal{D}f(y_1,y_2)dy_1dy_2 \\
        &\quad= \frac{1}{\sigma_2\sqrt{8\pi}}\sum_{m=1}^M D^m\int_0^1A^m(r)B^m(r)dr,
    }
    where
    \eqsnn{
        D^m &:= y_2^{m+1} - y_2^m, \\
        A^m(r) &:= \exp\parens{-\frac{1}{2\sigma_2^2}\brackets{(1-r)y_2^m + ry_2^{m+1} - \mu_2}^2}, \\
        B^m(r) &:= \erf\parens{\frac{(1-r)y_1^m+ry_1^{m+1} - \mu_1}{\sigma_1\sqrt{2}}}.
    }
    Finally, noting that $\mathbb{P}[Z\in\mathcal{D}]=\iint_\mathcal{D}f(y_1,y_2)dy_1dy_2$ completes the proof.
\end{proof}

\subsection{Efficiently Solving Batches of VLPs}
Program \eqref{opt:vlp} suggests solving a batch of $n_f \cdot n_v$ linear programs in parallel to compute the scaling values $\theta^{i,k}\in\R$. Here, we show that it is equivalent to further parallelize the LP computation in the following way. 

\begin{proposition}[Efficient VLP Batching]
    The following equality holds:
    \eqs{
        \theta^{i,k} = \min_{j=1,\dots,n_s}\theta^{i,k}_j,
    }
    where (with a slight abuse of notation) $\theta^{i,k}_j$ is the optimal solution to the following LP for a fixed index triple $(i,j,k)$.
    \begin{subequations}
    \label{opt:vlp_efficient}
    \begin{align}
        \maximize_{\theta^{i,k}_j \in \R, \; \alpha^{i,k}_j \in\R^{n_w}} \quad& \theta^{i,k}_j \\
        \subto \quad & \theta^{i,k} \geq 0 \\
        & \alpha^{i,k}_j \succeq 0 \\
        & \mathds{1}^\top \alpha^{i,k}_j = 1 \\
        & \parens{\theta^{i,k} d^{i,k}} T_j^i = -\widebar{W}\alpha^{i,k}_j.
    \end{align}
    \end{subequations}
\end{proposition}
\begin{proof}
    Follows by inspecting the dual programs.
\end{proof}

To actually solve many LPs in a batched manner, we use a custom port of the \verb|quantecon| implementation of the simplex method, available at the following link: \href{https://github.com/alberthli/jax_simplex}{github.com/alberthli/jax\_simplex}. Because this implementation is in \verb|JAX|, it can be run on both CPU or GPU without any additional modification. Due to certain parts of our computation stack being CPU-bound, we choose to compute the bound serially entirely on CPU. We observed that an \verb|Intel i9-12900KS| CPU typically exhibited about a 20\% increase in speed over an \verb|AMD Ryzen Threadripper PRO 5995WX|, which we attribute to speedups in Intel vs. ARM architectures on BLAS routines.

\subsection{Differentiating the PFC Bound}
In order to maximize the bound in \eqref{eqn:lfc} in a gradient-based nonlinear optimization program, we must compute the gradient of $L_{\text{fc}}$ with respect to the robot configuration $q$. To accomplish this, we need three major components:
\begin{itemize}
    \item differentiating through the numerical integration scheme used to evaluate the expressions in Proposition \ref{prop:bigauss} with respect to the polygon vertices $v^{i,k}$;
    \item differentiating through VLP$^{i,k}$ in \eqref{opt:vlp} (or the more efficient program \eqref{opt:vlp_efficient}) with respect to the wrench matrix $W(q)$;
    \item differentiating through the parameters $W$, $\bar{n}^i$, $\{\bar{t}^i_1,\bar{t}^i_2\}$, and $\{\sigma^i_1, \sigma^i_1\}$ with respect to the configuration $q$.
\end{itemize}

To differentiate through the numerical integration, we use the open source package \verb|torchquad| designed to differentiate numerical quadrature methods. All sub-expressions in Proposition \ref{prop:bigauss} can be implemented in \verb|JAX|, which yields the desired gradient in a straightforward manner.

To differentiate the optimal value of $\delta^{i,k}$ in \eqref{opt:vlp_efficient} with respect to the robot configuration $q$, we use implicit differentiation of the KKT conditions. The analytical gradient can be computed quickly in a nearly identical way to the one used by FRoGGeR, since here we also solve a linear program \cite[Prop. 1]{li2023_frogger}. For the the case of quadratic programs and more general programs, see \citeapp{amos2017_diffopt}.

We compute the gradients with respect to the uncertainty distribution parameters directly using \verb|JAX|. However, due to the special structure of \eqref{opt:vlp_efficient}, we compute the gradients of $W$ with respect to the distribution parameters completely analytically, which in practice leads to a large speedup. Because deriving these gradients is extremely tedious and involves an inordinate amount of algebraic manipulation, we defer the details to the open-source implementation, along with test code which verifies the correctness of our analytical derivations against the automatically-computed gradients from \verb|JAX|.

\subsection{Details for the Planar Integration}
All search directions $d^{i,k}$ are expressed in the local planar coordinates defined by the principal axes of the shape operator in Sec. \ref{sec:application}, which immediately yields a diagonal covariance matrix and planar points $v^{i,k}$. In practice, we do not integrate over the points generating the hull. Instead, we integrate over the polygon formed by connecting the points in lines counterclockwise, which may be smaller than the convex hull (for example, in Fig. \ref{fig:vertices}(b), see the point in the bottom left closest to the origin). We could easily quickly check which points lie on the hull boundary and only integrate over those - we choose not to for simplicity of implementation.

\subsection{Additional Experiment Details}
The main difference between our simulated and hardware experiments was the procedure in constructing the object model. In simulation, we found that learning neural SDFs using \verb|sdfstudio| \citeapp{yu2022_sdfstudio} was sufficient, because we could carefully control the training data by placing cameras around the object in a sphere. However, upon trying to recover a representation on real hardware, we were unable to reliably generate a coherent surface model.

Therefore, for the hardware experiments, we instead trained a neural radiance field (NeRF) \cite{mildenhall2020_nerf, nerfstudio} and represented the object surface as an empirically-chosen level set of the learned density function (in contrast with neural SDFs, where the object surface is always the 0-level set). In practice, we found that the quality of the object representation was sensitive to the choice of density level. It was easy to slightly over or undershoot an appropriate value, which led to grasping failures due to the belief that the object was bigger or smaller than in reality.

The picking trajectory optimization was done entirely in \verb|Drake| \cite{drake}, including collision avoidance constraints. In order to generate the requisite collision geometries associated with the object, we executed the marching cubes algorithm on the chosen density level set to recover a nonconvex mesh, which we then decomposed into convex bodies using VHACD.

\subsection{Acknowledgments}
We thank Lizhi (Gary) Yang for help setting up the hardware experiments. We thank Victor Dorobantu for discussions that led to a valid formulation and proof of Proposition \ref{prop:containment}. We thank Thomas Lew for discussions regarding Hausdorff distance bounds that inspired the condition of Proposition \ref{prop:containment}. We thank Georgia Gkioxari for discussions about object representations.

Finally, we thank the maintainers of all the open-source software used extensively in this work, including but not limited to \verb|Drake|, \verb|Pinocchio|, \verb|torchquad|, \verb|JAX|, \verb|sdfstudio|, \verb|trimesh|, \verb|nlopt|, \verb|VHACD|, and \verb|quantecon| \citeapp{gomez2021_torchquad, jax2018_github, yu2022_sdfstudio, trimesh, johnson2011_nlopt, kraft1988_slsqp}.

\bibliographystyleapp{unsrt}
\bibliographyapp{references}

\end{document}